\documentclass{isprs} 
\usepackage{subfigure}
\usepackage{setspace}
\usepackage{geometry} 
\usepackage{epstopdf}
\usepackage{epstopdf}
\usepackage{amsmath, amssymb}
\usepackage{graphicx}
\usepackage[labelsep=period]{caption}  
\usepackage[british]{babel} 
\usepackage[hang]{footmisc}
\usepackage{dblfloatfix}
\usepackage[colorlinks=true, linkcolor=blue, citecolor=blue, urlcolor=blue, breaklinks=true]{hyperref}
\usepackage{booktabs,caption}
\usepackage{ragged2e}
\usepackage{enumitem}

\begin{document}

\title{
From Propagation to Prediction: Point-level Uncertainty Evaluation of MLS Point Clouds under Limited Ground Truth
}

\author{Ziyang Xu\textsuperscript{1}, Olaf Wysocki\textsuperscript{3}, Christoph Holst\textsuperscript{1,2} }

\address{%
\centering
\parbox{\textwidth}{\centering
\textsuperscript{1} Chair of Engineering Geodesy, TUM School of Engineering and Design, Technical University of Munich, Munich, Germany — \{ziyang.xu, christoph.holst\}@tum.de\\[2pt]
\textsuperscript{2} TUM Leonhard Obermeyer Center, Technical University of Munich, Munich, Germany\\
\textsuperscript{3} CV4DT, University of Cambridge, Cambridgeshire CB3 0FA, Cambridge, UK — {okw24@cam.ac.uk}
}}

\abstract{
Evaluating uncertainty is critical for reliable use of Mobile Laser Scanning (MLS) point clouds in many high-precision applications such as Scan-to-BIM, deformation analysis, and 3D modeling. However, obtaining the ground truth (GT) for evaluation is often costly and infeasible in many real-world applications. To reduce this long-standing reliance on GT in uncertainty evaluation research, this study presents a learning-based framework for MLS point clouds that integrates optimal neighborhood estimation with geometric feature extraction. Experiments on a real-world dataset show that the proposed framework is feasible and the XGBoost model delivers fully comparable accuracy to Random Forest while achieving substantially higher efficiency (about 3 times faster), providing initial evidence that geometric features can be used to predict point-level uncertainty quantified by the C2C distance. In summary, this study shows that MLS point clouds' uncertainty is learnable, offering a novel learning-based viewpoint towards uncertainty evaluation research.
}

\keywords{Point Cloud Quality Check, Error Quantification, Machine Learning, Geometric Features, Uncertainty Regression.}
\maketitle

\section{Introduction}\label{Introduction}
\sloppy
Mobile Laser Scanning (MLS) systems are widely used to acquire 3D point clouds across different fields such as Architecture, Engineering, and Construction (AEC), engineering geodesy, and Geographic Information Systems (GIS) \cite{Xue2020}. Downstream tasks based on these point clouds, including registration, Scan-to-BIM, deformation analysis, and 3D modeling, require not only accurate geometry but also reliable estimates of uncertainty \cite{Geissendoerfer2025,xu2025c}. Moreover, errors in point clouds will inevitably accumulate and propagate through downstream workflows, making uncertainty evaluation indispensable \cite{Xu2025b}.

Traditional point clouds uncertainty evaluation methods can be grouped into forward and backward modeling \cite{Shi2021}. Forward modeling relies on variance and covariance propagation and detailed theoretical analysis of involved errors. It models the various error sources to infer the uncertainty of the final point clouds. In MLS systems, four main categories of error sources contribute to uncertainty: instrumental errors, atmospheric errors, object- and geometry-related errors, and trajectory estimation errors \cite{Habib2009,Schenk2001}. However, MLS point clouds' uncertainty is also influenced by many other factors, including the observation environment, internal SLAM algorithms, and scanning speed. As a result, accurate and comprehensive modeling of all error sources is rarely feasible \cite{Holst2016,Shi2021}.

In contrast, backward modeling, which is the more common solution, avoids explicit error modeling and propagation. Instead, it uses accurate reference data to evaluate uncertainty empirically. Its main advantage is that uncertainty is quantified from actual measurements, which supports evaluation procedures that generalize across different systems. The core of backward modeling is to establish reliable correspondences between the scanned and the reference data to characterize their discrepancies.

Although backward modeling is more applicable than forward modeling, a crucial concern remains. Most backward solutions assume access to high-precision reference data as the ground truth (GT), for example, Terrestrial Laser Scanning (TLS) or high-fidelity as-built models, and then use the GT to evaluate MLS point clouds \cite{xu2025}. This assumption is theoretically sound, but in practice, users often face three limitations: 

\begin{enumerate}[label=(\arabic*), itemsep=0em, parsep=0em, topsep=0em, partopsep=0em]
\item Evaluation depends heavily on the GT and will be problematic if the GT is unavailable, which is frequently the case in real-world applications.
\item Obtaining the GT often costs much more than common data collection in terms of money and time, which limits adoption.
\item Inherent practical paradox, namely that if the GT is already available, the value of performing MLS data collection for evaluation becomes questionable.
\end{enumerate}

To overcome these limitations, learning-based solutions have begun to gain attention \cite{hartmann2023a,hartmann2023b}. The core idea is to recast error quantification as a supervised learning problem, in which a model learns the relationship between each point’s error and its local geometric or attribute features. Unlike statistical error propagation or fully GT-based evaluations, this data-driven paradigm requires reference data as GT only during training; once trained, the model no longer depends on it. This design improves flexibility and offers substantial potential for practical deployment. In this context, this study proposes a framework of point-level uncertainty evaluation based on machine learning under limited GT. The main contributions are as follows:

\begin{itemize}
\setlength\itemsep{0em}\setlength\parskip{0em}\setlength\topsep{0em}\setlength\partopsep{0em}\setlength\parsep{0em} 
\item{A learning-based framework is designed to predict point-level uncertainty from its geometric features for MLS point clouds.} 

\item{A controlled comparison of XGBoost and Random Forest implementation is conducted, showing comparable accuracy, with XGBoost achieving lower runtime.}

\item{A feature importance analysis reveals how important features correspond to interpretable geometric patterns that drive error growth.}
\end{itemize}

\section{Related Works}\label{Related Works}

\subsection{Forward Modeling}\label{Forward Modeling}

Forward modeling is based on the variance-covariance propagation law. A detailed theoretical analysis of all the errors of laser scanning systems is presented, in which all kinds of error sources are modeled, and the error characteristics of the final point clouds are obtained. Several forward modeling-based evaluations are realized, considering the error information from the different sensors \cite{Barber2007,Goel2015,Xu2016}. Some studies aim to establish error budgets for controlling point clouds quality \cite{Habib2009}. The ranging precision, based on the intensity information of the point clouds, has also been explored \cite{Stenz2020,Wujanz2017}. As previously mentioned, forward modeling is theoretically feasible but faces several limitations. Consequently, an increasing number of studies are shifting toward backward modeling.

\subsection{Backward Modeling }\label{Backward Modeling}

Backward modeling relies on accurate reference data as the GT to quantify errors empirically. To achieve this, the fundamental procedure involves accurately and efficiently quantifying predefined uncertainty metrics with respect to the GT. Standard practices compute cloud-to-cloud (C2C), cloud-to-mesh (C2M), and multiscale model-to-model cloud comparison (M3C2) distances to measure discrepancies between scans and the GT \cite{Lague2013,Heinz2015,Stenz2017,Zahs2022,Xu2025b}. 
These distance-based evaluations offer simplicity and low cost, but they lack 6 degrees of freedom (DoF) transformation parameters, so they cannot comprehensively represent uncertainty from a 3D perspective. To handle this gap, point-based and line-based methods have been introduced \cite{Poreba2013,Hauser2016}. Plane-based strategies are also widely used; they exploit planar primitives to establish robust correspondences and to evaluate uncertainty \cite{Shi2021,Shahraji2020}. The key is using geometric features of point clouds to establish correspondences, followed by quantifying uncertainty estimates for 6 DoF. In the context of MLS, this is often used to evaluate trajectory estimation errors \cite{xu2025}.

Owing to its flexibility and ease of use, backward modeling remains the predominant solution. Its reliance on the GT ensures its reliability and accuracy, but also limits scalability to large real-world deployments. Learning-based modeling has therefore emerged as a promising direction.

\subsection{Learning-based Modeling}\label{Learning-based Modeling}

Learning-based uncertainty evaluation is an emerging paradigm in the context of geodesy. It stems from attempts and explorations to reduce or eliminate dependence on the GT using learning techniques. Initial idea used ensemble methods to estimate ranging precision in TLS from features such as intensity and incidence angle \cite{hartmann2023a,hartmann2023b}. Subsequent work extended this idea to deep learning to predict and calibrate TLS ranging accuracy \cite{hartmann2024}. These investigations mainly focused on TLS distance measurement under laboratory conditions and did not address MLS point clouds. More recently, ensemble learning has been explored for MLS uncertainty by framing the task as a binary classification of point quality in a real-world scenario \cite{xu2025d}.

To the best of our knowledge, most existing research on evaluating the uncertainty of point clouds primarily centers on forward and backward modeling. Only a limited number of publications utilize machine learning for this purpose on TLS, indicating that this area is still in its early stages of development. This study presents the first learning-based uncertainty modeling applied to a real-world MLS dataset.

\section{Methodology}\label{Methodology}

\begin{figure*}[!hbt]
\centering
\includegraphics[width=0.95\textwidth]{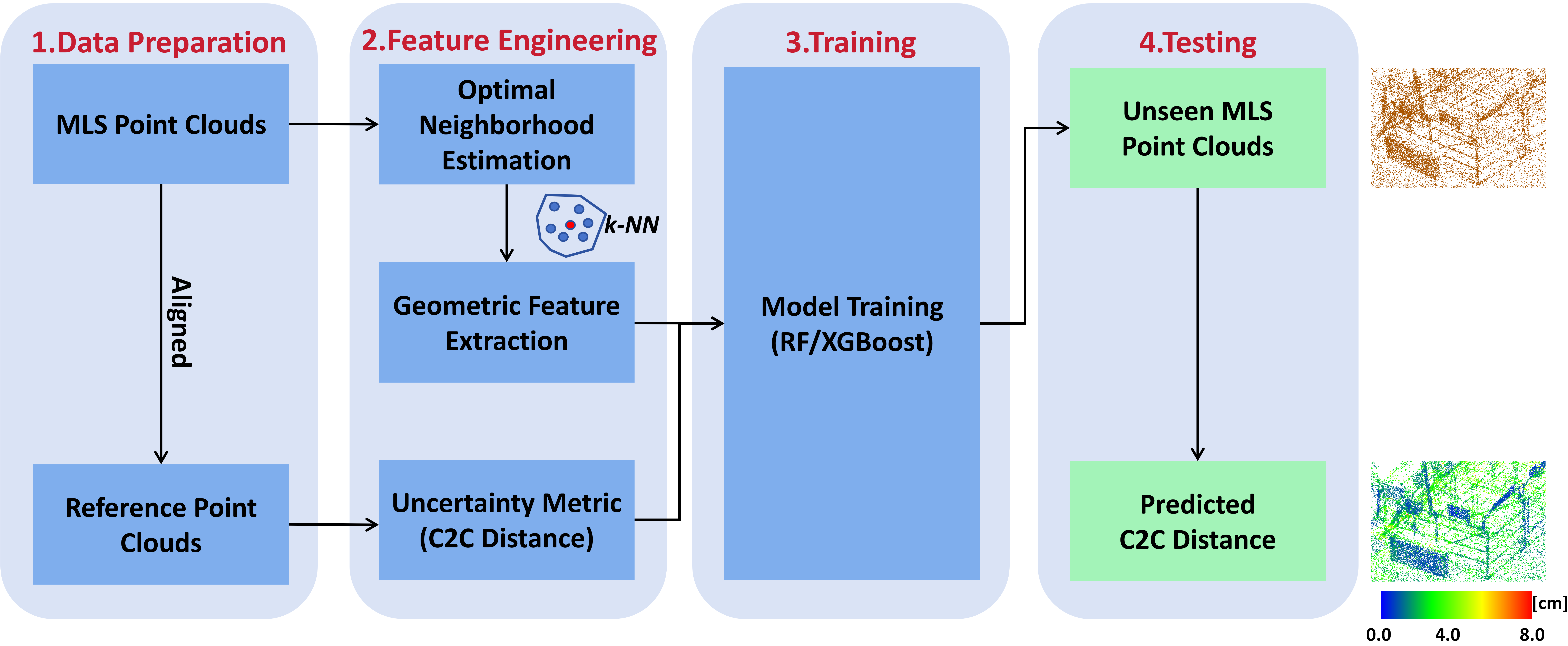}
\caption{Proposed learning-based point-level uncertainty evaluation framework.}
\label{fig:workflow}
\end{figure*}

This section presents the proposed framework as shown in Figure \ref{fig:workflow}. The framework starts from data preparation in Section \ref{Data Preparation}, then covers a description of feature engineering in Section \ref{Feature Engineering}. After that, more information about model training and testing is presented in Section \ref{Training} and Section \ref{Testing}, respectively.

\subsection{Data Preparation}\label{Data Preparation}

As seen in the framework, two inputs are MLS point clouds and the corresponding reference point clouds, preferably obtained via TLS. In this study, point-level uncertainty is quantified using the C2C distance. C2C distance is preferred over M3C2 distance for two reasons. First, C2C distance is more responsive to noise at the individual point level and therefore provides a more direct representation of local error characteristics. Second, M3C2 distance requires a user-defined radius, and its results are sensitive to that choice, whereas C2C distance does not involve such a scale parameter. 

Since this framework aims to learn a mapping from point-level geometric features to the corresponding uncertainty metric (quantified by C2C distance). To reduce label noise caused by density differences between MLS and TLS data and by residual minor alignment errors (less than 10 mm), only MLS points with a C2C distance less than 80 mm are retained for model training. 

\subsection{Feature Engineering}\label{Feature Engineering}

As aforementioned, this study aims is to examine whether point-level C2C distance is related to local geometric features. In other words, it investigates whether a point’s error behavior can be predicted and interpreted from its geometric feature domain, which makes point-level feature extraction a critical step. Moreover, only with a suitable neighborhood, local geometric features can be estimated reliably at the point level. Selecting this neighborhood remains a fundamental challenge, since different definitions can introduce additional variability into the extracted features.

It has been proven that neighborhoods defined by the $k$-nearest neighbors ($k$-NN) offer clear advantages, and the geometric features derived from them can improve point cloud classification performance \cite{weinmann2015}. Motivated by this evidence, this study adopts the same optimal neighborhood estimation strategy. For each point in the input MLS point clouds, the optimal number of neighbors $OptN$ is first estimated. Based on $OptN$, the corresponding 3D and 2D geometric features are then computed. In total, 26 features will be used for training to predict uncertainty. Additionally, these geometric features have also been used for other 3D learning tasks to achieve better performance, such as classification and segmentation \cite{tan2023,robert2023,wysocki2023}. Table \ref{tab:geom_feats_all} lists basic definitions of these geometric features. Detailed calculations are outside the scope of this study. Please refer to these publications \cite{weinmann2014,weinmann2015}.

\begin{table*}[h!]
\centering
\small
\resizebox{\linewidth}{!}{%
\begin{tabular}{r l l c l l}
\toprule
\textbf{\#} & \textbf{Feature} & \textbf{Formula / Definition} & \textbf{Range} & \textbf{Unit} & \textbf{Notes} \\
\midrule
1  & $linearity$ & $(l_1-l_2)/l_1$ & $[0,1]$ & -- & line-like strength \\
2  & $planarity$ & $(l_2-l_3)/l_1$ & $[0,1]$ & -- & plane-like strength \\
3  & $scattering$ & $l_3/l_1$ & $[0,1]$ & -- & $\lambda_3/\lambda_1$ \\
4  & $omnivariance$ & $(l_1l_2l_3)^{1/3}$ & $[0,1/3]$ & -- & geometric mean of normalized EVs \\
5  & $anisotropy$ & $(l_1-l_3)/l_1$ & $[0,1]$ & -- & directionality contrast \\
6  & $eigenentropy$ & $-\sum_{i=1}^3 l_i\ln l_i$ & $[0,\ln 3]$ & -- & larger if more isotropic \\
7  & $sum\_EVs$ & $\lambda_1+\lambda_2+\lambda_3$ & $\ge 0$ & m$^2$ & total 3D variance \\
8  & $change\_of\_curvature$ & $l_3$ & $[0,1]$ & -- & noise-sensitive \\
9  & $Z\_vals$ & $z$ & -- & m & absolute height \\
10 & $radius\_kNN$ & $r_{3D}$ to the $k$-th neighbor & $\ge 0$ & m & neighborhood scale (3D) \\
11 & $density$ & $\dfrac{k+1}{\tfrac{4}{3}\pi r_{3D}^3}$ & $\ge 0$ & points/m$^3$ & volumetric density \\
12 & $verticality$ & $1 - v_z$ (eigenvector of $\lambda_1$) & $\approx[0,2]$ & -- & uses major axis, not $1-|n_z|$ \\
13 & $delta\_Z\_kNN$ & $z_{\max}-z_{\min}$ in $k$-NN & $\ge 0$ & m & vertical relief in $k$-NN \\
14 & $std\_Z\_kNN$ & unbiased $\sigma_z$ in $k$-NN & $\ge 0$ & m & vertical roughness \\
15 & $radius\_kNN\_2D$ & $r_{2D}$: max XY dist among first $k{+}1$ & $\ge 0$ & m & projected $k$-NN radius \\
16 & $density\_2D$ & $\dfrac{k+1}{\pi r_{2D}^2}$ & $\ge 0$ & points/m$^2$ & planar density \\
17 & $sum\_EVs\_2D$ & $\mu_1+\mu_2$ & $\ge 0$ & m$^2$ & total variance in XY \\
18 & $EV\_ratio$ & $m_2/m_1=\mu_2/\mu_1$ & $[0,1]$ & -- & XY anisotropy; smaller $\rightarrow$ linear \\
19 & $frequency\_acc\_map$ & points per XY grid cell & $\ge 0$ & -- & global 2D grid \\
20 & $delta\_z$ & $z_{\max}-z_{\min}$ per XY cell & $\ge 0$ & m & global grid relief \\
21 & $std\_z$ & sample $\sigma_z$ per XY cell & $\ge 0$ & m & global grid roughness \\
22 & $EV3D\_1$ & $l_1=\lambda_1/(\sum\lambda)$ & $[0,1]$ & -- & largest normalized 3D EV \\
23 & $EV3D\_2$ & $l_2=\lambda_2/(\sum\lambda)$ & $[0,1]$ & -- & middle normalized 3D EV \\
24 & $EV3D\_3$ & $l_3=\lambda_3/(\sum\lambda)$ & $[0,1]$ & -- & smallest normalized 3D EV \\
25 & $EV2D\_1$ & $m_1=\mu_1/(\mu_1+\mu_2)$ & $[0,1]$ & -- & largest normalized 2D EV \\
26 & $EV2D\_2$ & $m_2=\mu_2/(\mu_1+\mu_2)$ & $[0,1]$ & -- & smallest normalized 2D EV \\
\bottomrule
\end{tabular}
}
\caption*{\justifying \textbf{Notes:} 
(1) $k$ is selected per point by entropy minimization: $H(k)=-\sum_i p_i\ln p_i$ with $p_i=\lambda_i/(\lambda_1+\lambda_2+\lambda_3)$; pick $k=\arg\min H(k)$. The $k$-NN includes the query point, thus $n=k+1$. 
(2) Unbiased covariance is used when $n>1$ (scaled by $n/(n-1)$). 3D eigenvalues (EV) are sorted as $\lambda_1\ge\lambda_2\ge\lambda_3>0$ and normalized $l_i=\lambda_i/(\lambda_1+\lambda_2+\lambda_3)$. For 2D (XY), $\mu_1\ge\mu_2>0$ and $m_i=\mu_i/(\mu_1+\mu_2)$. 
(3) \textit{Verticality} uses the major eigenvector ($\lambda_1$); a more “surface-normal” variant would use $1-|n_z|$ with $n$ the eigenvector of $\lambda_3$. 
(4) $r_{3D}$ is the distance to the $k$-th neighbor in 3D; $r_{2D}$ is the maximum XY distance among the first $k{+}1$ neighbors (ordered by 3D distance). 
(5) Grid features (19--21) are computed on a global XY raster with a given cell size and are independent of $k$.
}
\caption{Basic definitions and explanations of 26 point-level geometric features.}
\label{tab:geom_feats_all}
\end{table*}

\subsection{Training}\label{Training}

After geometric feature extraction, each point $P_i$ is represented by a 27-dimensional feature vector $\mathbf{x}_i \in \mathbb{R}^{27}$ that concatenates its 26 geometric features and the corresponding optimal neighborhood size $OptN_i$. Stacking all samples yields the design matrix $\mathbf{X} = [\mathbf{x}_1,\ldots,\mathbf{x}_n]^\top \in \mathbb{R}^{n\times 27}$ and the target vector $\mathbf{y} = (y_1,\ldots,y_n)^\top \in \mathbb{R}^{n}$, where $y_i$ denotes the C2C distance of point $P_i$. The task is formulated as supervised regression, seeking a function $f:\mathbb{R}^{27}\!\to\!\mathbb{R}$ such that $\hat{\mathbf{y}} = f(\mathbf{X})$ approximates $\mathbf{y}$. Two commonly used ensemble models, Random Forest (RF) \cite{breiman2001} and XGBoost \cite{chen2016}, are employed to obtain the predictions $\hat{\mathbf{y}}$.

RF and XGBoost are selected for two reasons. From a technical standpoint, both are tree-ensemble regressors that capture nonlinear relationships and feature interactions, handle mixed-scale inputs with minimal preprocessing, and are robust to outliers. RF reduces variance through bootstrap aggregation and random feature subspaces, whereas XGBoost improves accuracy and efficiency through gradient boosting with regularization and optional GPU acceleration. Both also support model attribution, using permutation importance for RF and SHAP values for XGBoost. From an empirical standpoint, prior studies report promising results with RF and XGBoost on TLS measurement \cite{hartmann2023a,hartmann2023b}.

\begin{table}[t]
  \centering\small
  \resizebox{0.95\linewidth}{!}{%
    \begin{tabular}{lll}
      \toprule
      \textbf{Hyperparameter} & \textbf{Value} & \textbf{Notes} \\
      \midrule
      \texttt{n\_estimators} & 100 & Number of trees \\
      \texttt{max\_depth}    & 20  & Depth cap \\
      \texttt{max\_samples}  & 0.5 & Subsample per tree \\
      \texttt{class\_weight} & Balanced & Handles class imbalance \\
      \texttt{n\_jobs}       & 1   & Single-thread \\
      \texttt{random\_state} & Random seed & Reproducibility \\
      \bottomrule
    \end{tabular}%
  }
  \caption{Key Random Forest hyperparameters.}
  \label{tab:rf_hparams_cls}
\end{table}

\begin{table}[t]
  \centering\small
  \resizebox{\linewidth}{!}{%
    \begin{tabular}{lll}
      \toprule
      \textbf{Hyperparameter} & \textbf{Value} & \textbf{Notes} \\
      \midrule
      \texttt{max\_depth}               & 8            & Tree depth \\
      \texttt{eta}                      & 0.05         & Learning rate \\
      \texttt{subsample}                & 0.8          & Row sampling per round \\
      \texttt{colsample\_bytree}        & 0.8          & Feature sampling per tree \\
      \texttt{num\_boost\_round}        & 1000         & Maximum boosting rounds \\
      \texttt{early\_stopping\_rounds}  & 50           & Patience on validation loss \\
      \texttt{seed}                     & Random seed  & Reproducibility \\
      \bottomrule
    \end{tabular}%
  }
  \caption{Key XGBoost hyperparameters.}
  \label{tab:xgb_hparams_cls}
\end{table}

To clarify the training protocol, Tables~\ref{tab:rf_hparams_cls} and~\ref{tab:xgb_hparams_cls} list the key hyperparameters used in this study. The XGBoost setup employs histogram-based boosting with GPU execution and early stopping. The Random Forest configuration adopts memory-safe settings on the CPU. Additionally, a 5-fold grid-based strategy was employed to mitigate data leakage. Groups are formed on a regular spatial grid (grid size = 3 m). For each fold, the model is trained on four folds and validated on the held-out fold. The averaged results across the five folds were reported to provide a comprehensive and spatially unbiased evaluation of model performance.

\subsection{Testing}\label{Testing}

The regression performance is tested with the aforementioned 5-fold cross-validation strategy using five complementary metrics: root mean squared error (RMSE), 
\begin{equation}
\mathrm{RMSE}=\sqrt{\frac{1}{n}\sum_{i=1}^{n}\bigl(y_i-\hat{y}_i\bigr)^2},
\end{equation}
emphasizing larger errors due to the quadratic penalty; mean absolute error (MAE), 
\begin{equation}
\mathrm{MAE}=\frac{1}{n}\sum_{i=1}^{n}\bigl|y_i-\hat{y}_i\bigr|,
\end{equation}
reflecting the average absolute deviation and is less sensitive to outliers than RMSE; median absolute error (MedAE), 
\begin{equation}
\mathrm{MedAE}=\operatorname{median}_{i}\bigl|y_i-\hat{y}_i\bigr|,
\end{equation}
providing a robust estimate of the typical absolute error; coefficient of determination ($R^2$), 
\begin{equation}
R^2 \;=\; 1-\frac{\sum_{i=1}^{n}\bigl(y_i-\hat{y}_i\bigr)^2}{\sum_{i=1}^{n}\bigl(y_i-\bar{y}\bigr)^2},
\qquad
\bar{y}=\frac{1}{n}\sum_{i=1}^{n} y_i,
\end{equation}
measuring the proportion of variance in $y$ explained by the predictions; and tolerance-aligned accuracy $\mathrm{P@}m$ for $m\in\{10,20,30,40,50\}$\,mm,
\begin{equation}
\mathrm{P@}m \;=\; \frac{1}{n}\sum_{i=1}^{n} \mathbf{1}\!\left\{\,\bigl|y_i-\hat{y}_i\bigr|\times 100 \le k\,\right\}\times 100\%,
\end{equation}
reporting the percentage of points whose absolute error does not exceed $m$\,mm, which aligns directly with common tolerance thresholds. All metrics are computed on the predicted values $\hat{y}_i$ and references $y_i$.

\section{Results and Analysis}\label{Results and Analysis}

This section introduces the experiment dataset in Section \ref{Experiment Dataset} and presents quantitative results in Section \ref{Quantitative Results}, then visualizes the predicted results in Section \ref{Visualized  Results}, and gives feature importance analysis in Section \ref{Feature Importance Analysis}.

\subsection{Experiment Dataset}\label{Experiment Dataset}

\begin{figure}[ht!]
\begin{center}
		\includegraphics[width=0.9\linewidth]{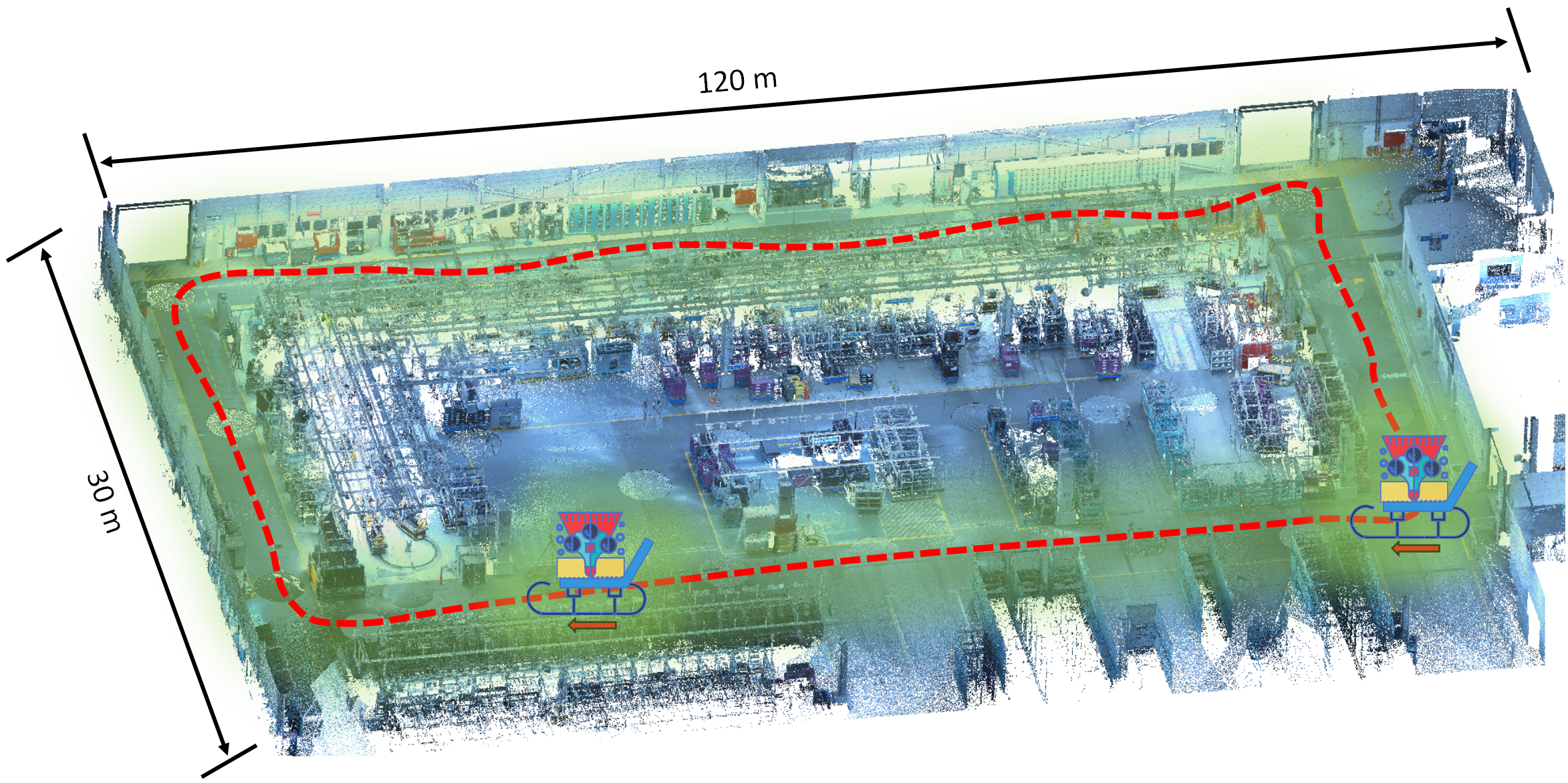}
	\caption{Overview of the scanned area. Red line indicates the scanning path.}
\label{fig:scanned area}
\end{center}
\end{figure}

All data was collected in an assembly hall at the BMW factory, an indoor industrial environment that combines complex geometries and includes both reflective and absorbent materials. An overview of the scanned area is shown in Figure \ref{fig:scanned area}. MLS point clouds were acquired from the Emesent Hovermap ST-X unit\footnote{\url{https://emesent.com/hovermap-series/}}. Reference point clouds were acquired with a Trimble X9 terrestrial laser scanner\footnote{\url{https://geospatial.trimble.com/en/products/hardware/trimble-x9}}. After this, using a high-precision total station control network and black \& white targets, two datasets are accurately aligned within the same coordinate system \cite{xu2025}. More information about the experiment scenario can be found in Table \ref{experiment scenario}.

\begin{table}[h]
\centering
\resizebox{0.9\linewidth}{!}{%
\begin{tabular}{ll}
\toprule
  Scanning Area & About 3,500 m$^{2}$ \\ 
  Number of Points & 5,000,000 \\
  MLS System & EMESENT HOVERMAP ST-X\ \\
  Accuracy Specification* & $\pm$15mm \\
  Number of black \& white targets & 12\\
 \bottomrule
  {\small * manufacturer's specification in typical environment.}\\
  \end{tabular}
}
\caption{Basic information on experiment scenario.}
\label{experiment scenario}
\end{table}

\subsection{Quantitative Results}\label{Quantitative Results}

Table \ref{tab:reg-core-metrics-vert} reports the core performance metrics averaged over five folds with 95\% confidence intervals. RF and XGBoost achieve virtually identical accuracy: both obtain an RMSE of 10.9\,mm; the MAE differs by 0.1\,mm; the MedAE is 4.0\,mm for both; and the $R^2$ values are close with overlapping intervals. The tolerance-aligned results in Table~\ref{tab:reg-threshold-accuracy-vert} are similarly close across all thresholds: RF is slightly higher at P@10-20\,mm, whereas XGBoost is slightly higher at P@40-50\,mm, and all differences lie within the reported confidence intervals. 

The principal distinction is computational efficiency: the average runtime per fold is 210.5\,s for RF and 74.6\,s for XGBoost. The efficiency advantage of XGBoost arises from both algorithmic design and GPU support. First, XGBoost employs histogram-based split finding with quantile binning, which reduces the number of candidate thresholds per feature and lowers computation and memory traffic. Second, XGBoost significantly elevates performance by harnessing the power of parallel processing, efficiently accumulating gradients and Hessians across multiple GPU threads. 

\begin{table}[ht]
  \centering
  \small
  \begin{tabular}{lcc}
    \toprule
    \textbf{Metric} & \textbf{RF} & \textbf{XGBoost} \\
    \midrule
    RMSE (mm) $\downarrow$   & $10.9 \pm 0.6$ & $10.9 \pm 0.6$ \\
    MAE (mm)  $\downarrow$   & $6.5 \pm 0.4$ & $6.6 \pm 0.3$ \\
    MedAE (mm) $\downarrow$  & $4.0 \pm 0.2$ & $4.0 \pm 0.2$ \\
    $R^2$ $\uparrow$ & $0.36 \pm 0.02$ & $0.37 \pm 0.03$ \\
    Runtime / fold (s) $\downarrow$ & 210.5 & \textbf{74.6} \\
    \bottomrule
  \end{tabular}
   \caption{Regression performance (mean $\pm$ 95\% CI).}
  \label{tab:reg-core-metrics-vert}
\end{table}

\begin{table}[ht]
  \centering
  \small
  \begin{tabular}{lcc}
    \toprule
                       & \textbf{RF} & \textbf{XGBoost} \\
    \midrule
    P@10\,mm  & $0.832 \pm 0.015$ & $0.823 \pm 0.013$ \\
    P@20\,mm  & $0.945 \pm 0.007$ & $0.941 \pm 0.007$ \\
    P@30\,mm  & $0.973 \pm 0.004$ & $0.973 \pm 0.004$ \\
    P@40\,mm  & $0.983 \pm 0.003$ & $0.985 \pm 0.003$ \\
    P@50\,m  & $0.990 \pm 0.002$ & $0.992 \pm 0.002$ \\
    \bottomrule
  \end{tabular}
  \caption{Proportion within absolute error thresholds (mean $\pm$ 95\% CI).}
  \label{tab:reg-threshold-accuracy-vert}
\end{table}

Overall, the two models provide fully comparable prediction accuracy, whereas XGBoost offers a clear efficiency advantage, making it a more feasible solution for practical application. Therefore, subsequent visualization results and feature importance analyses will focus primarily on XGBoost.

\subsection{Visualized  Results}\label{Visualized  Results}

\begin{figure}[h!]
  \centering
  \includegraphics[width=1.0\linewidth]{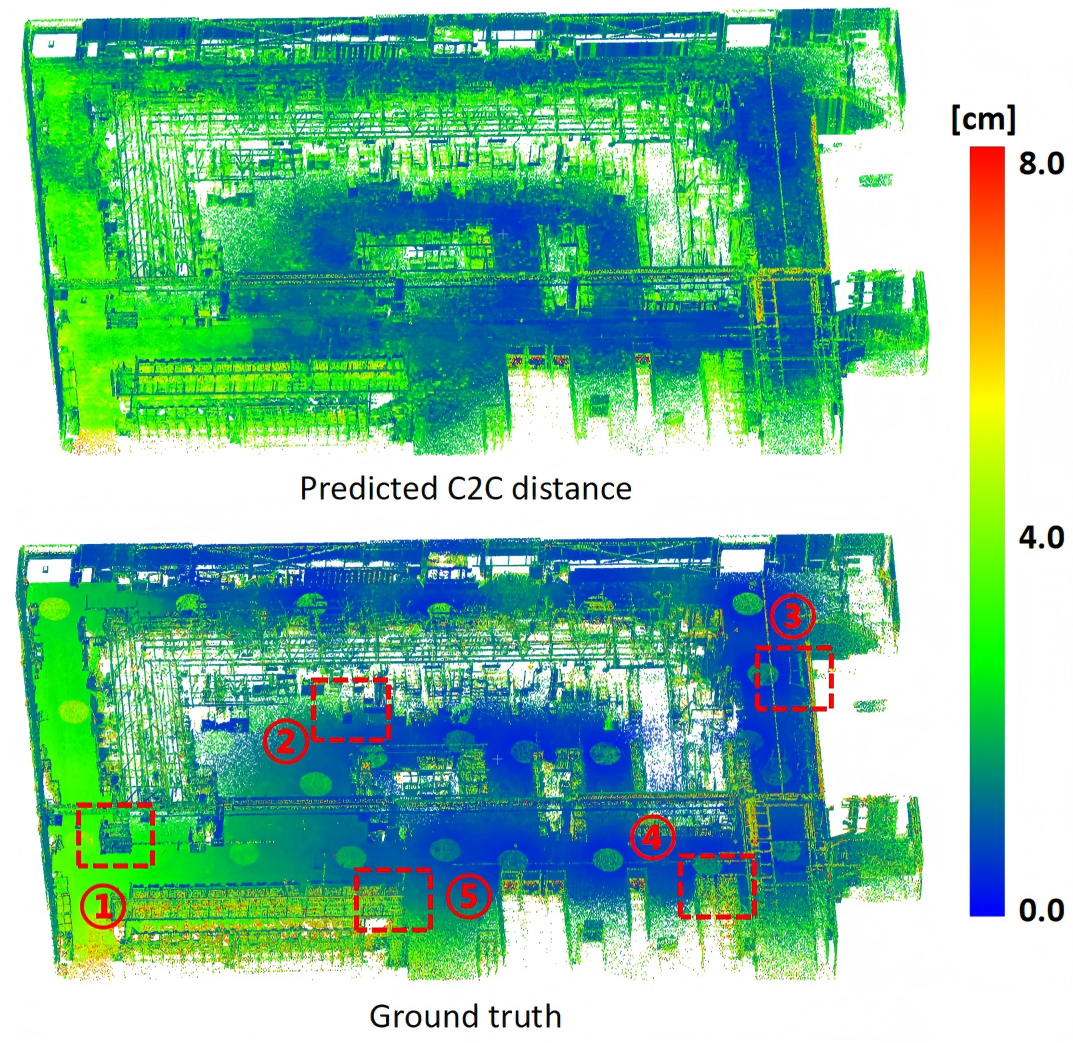}
  \caption{Predicted results from XGBoost model.}
  \label{fig:C2C pred}
\end{figure}

\begin{figure*}[h!]
  \centering
  \includegraphics[width=1.0\linewidth]{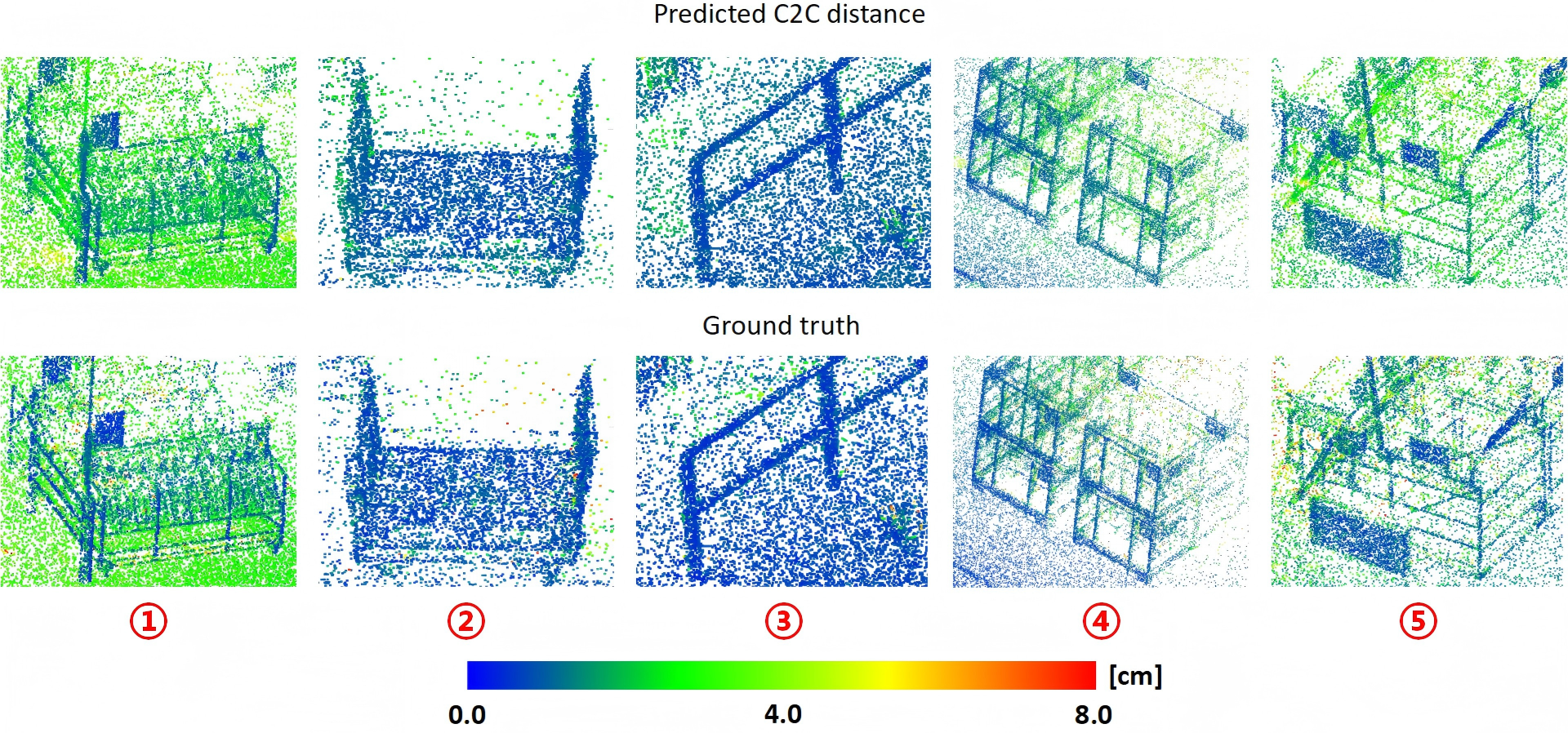}
  \caption{Detailed comparison of XGBoost prediction on selected objects.}
  \label{fig:C2C details}
\end{figure*}

Figure~\ref{fig:C2C pred} compares predicted C2C distances from XGBoost with the GT. Most of the predictions closely follow the GT, particularly on smooth and continuous areas. To examine local behavior, five objects with distinct geometric structures are selected from Figure~\ref{fig:C2C pred} and shown in detail in Figure~\ref{fig:C2C details}. Across these objects, it maintains fine-scale fidelity and achieves accurate estimates even in geometrically intricate areas.

\begin{figure*}[h]
  \centering
  \includegraphics[width=0.9\linewidth]{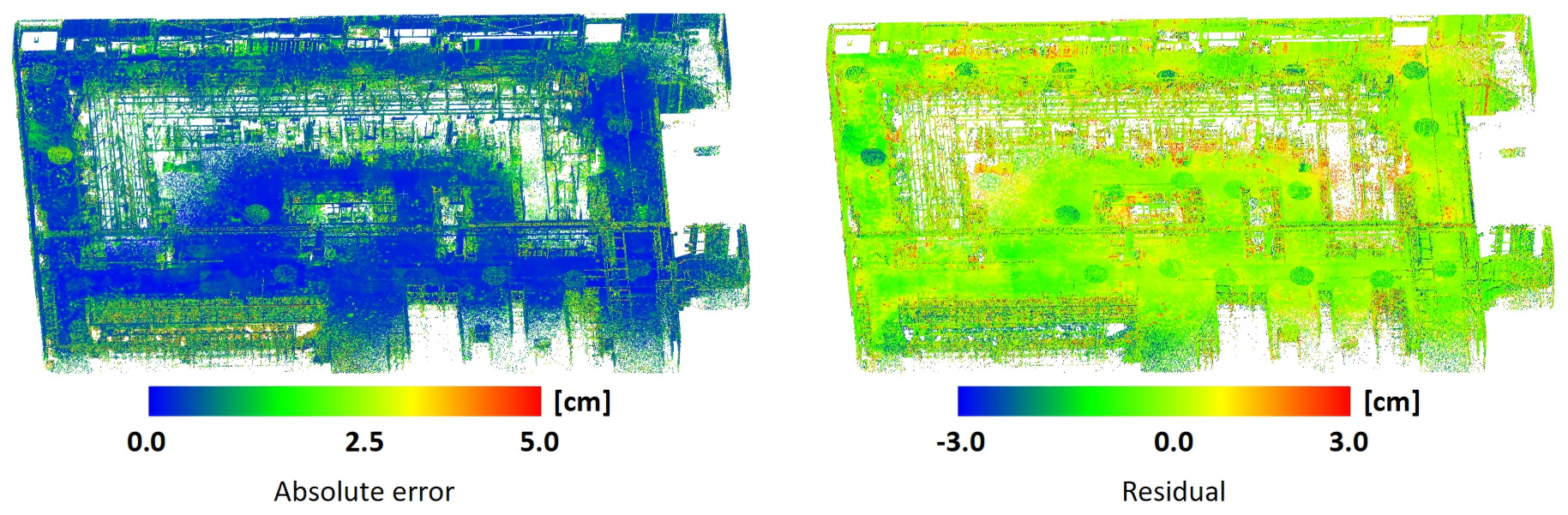}
  \caption{Absolute error and residual of the predicted C2C distance with respect to the GT from XGBoost model.}
  \label{fig:C2C error}
\end{figure*}

To further quantify and visualize the prediction performance of XGBoost, Figure~\ref{fig:C2C error} reports the absolute error $\lvert y-\hat{y}\rvert$ and the residual $r=\hat{y}-y$ of the predicted C2C distances with respect to the GT. The results corroborate the overall accuracy of the predictions, with minor errors across most areas. Increased variability is observed in areas with highly complex or sparse geometry, such as slender rod-like objects and thin linear features, where predictions appear less stable. This behavior is consistent with the known inherent sensitivity of the C2C distance calculation in such structures, which complicates both measurement and prediction.

\subsection{Feature Importance Analysis}\label{Feature Importance Analysis}


\paragraph{SHAP Feature Importance Analysis} \label{SHAP Feature Importance}

\begin{figure}[htb!]
  \centering
  \includegraphics[width=1.0\linewidth]{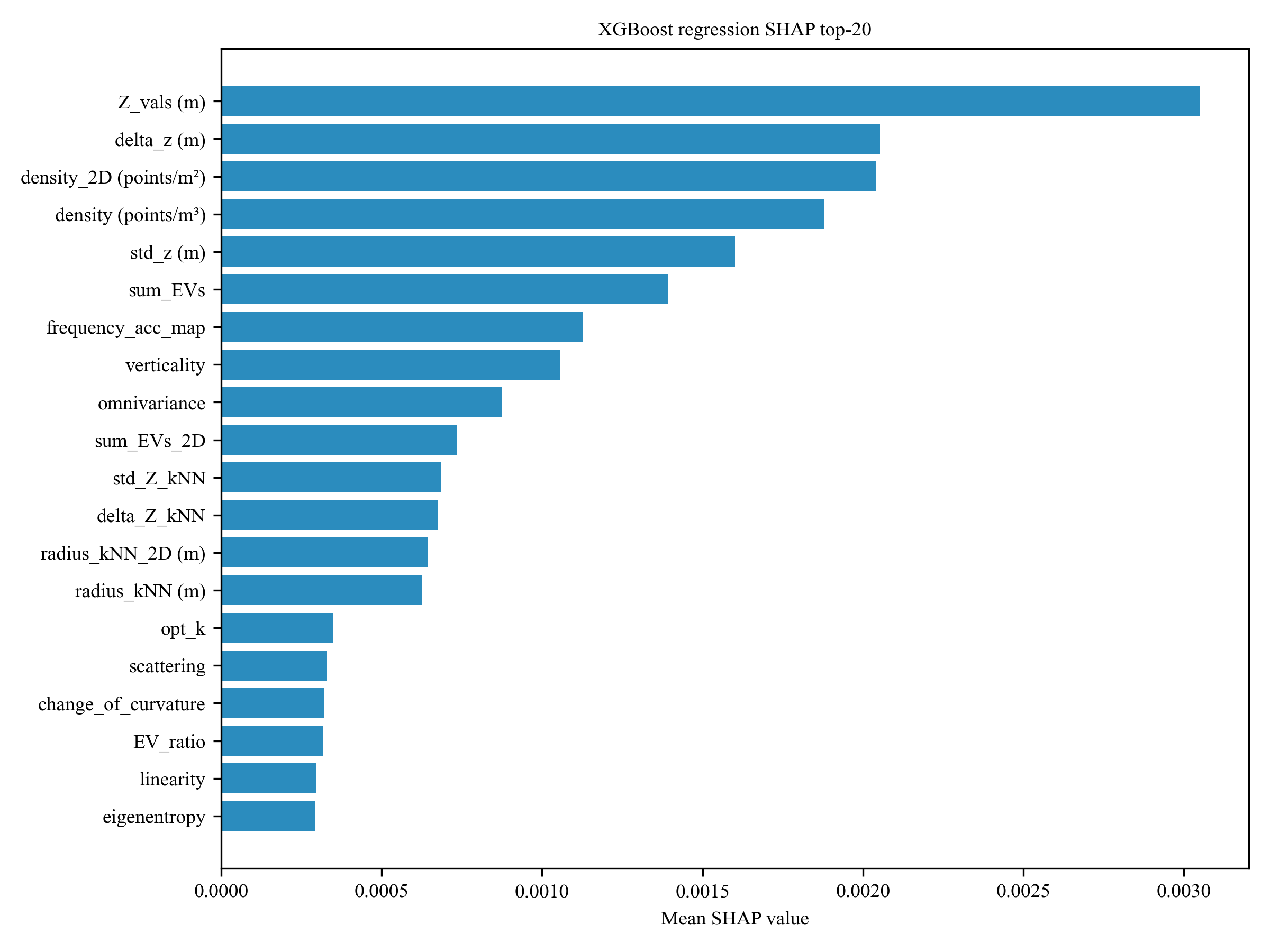}
  \caption{Top 20 most important features of XGBoost model.}
  \label{fig:feature_importance_top20}
\end{figure}

\begin{figure}[htb!]
  \centering
  \includegraphics[width=1.0\linewidth]{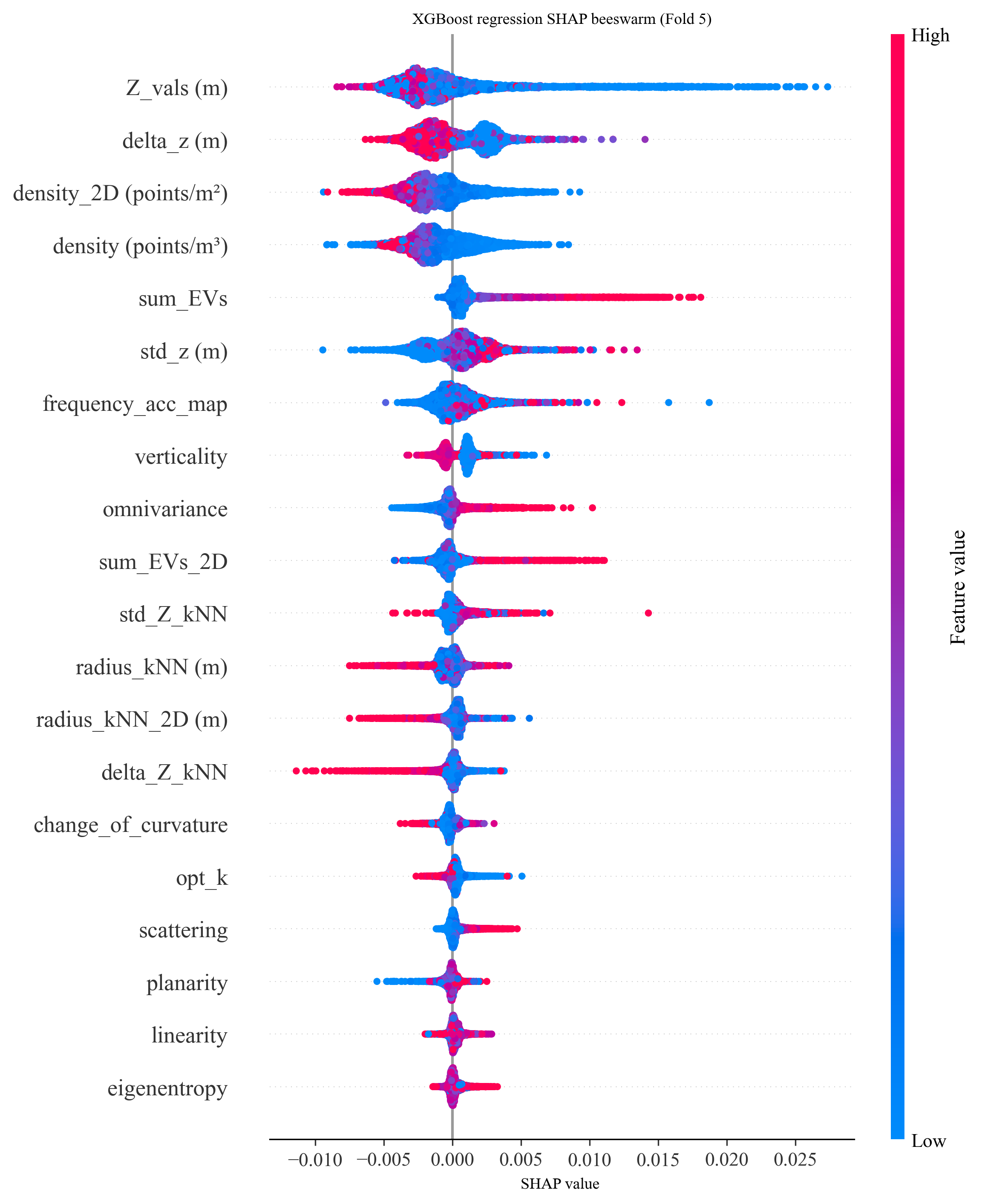}
  \caption{SHAP beeswarm of XGBoost model.}
  \label{fig:SHAP beeswarm}
\end{figure}

Figure \ref{fig:feature_importance_top20} summarizes global contributions by mean SHAP value, indicating that height-variation and sampling-density features dominate ($Z\_vals$, $delta\_z$, $std\_z$, $density\_2D$, $density$) XGBoost prediction. These are followed by eigenvalue-derived shape measures and orientation features, while neighborhood-scale features such as $radius\_kNN$ show moderate but consistent influence. Figure \ref{fig:SHAP beeswarm} indicates that the top four features have more red points concentrated on the left side of the horizontal axis, which means high feature values tend to decrease the predicted C2C distance. This pattern identifies height and sampling density as the two dominant factors. The underlying reasons are twofold. First, higher density usually reflects shorter range, a more favorable incidence angle, or slower platform motion, which stabilizes neighborhood statistics and reduces nearest-neighbor mismatches; therefore, higher density leads to lower predicted C2C distance. Second, in this dataset, there is no ceiling; the point clouds only consist of ground and surrounding structures, and most errors originate from the vertical direction \cite{xu2025}. Points at higher $Z\_vals$ therefore mainly belong to walls or elevated components that are more regularly shaped and better observed, which explains why larger $Z\_vals$ and, in this scene, larger local height variation $delta\_z$ are associated with lower predicted C2C distance.

In summary, the SHAP-based feature analysis of the XGBoost model is only intended to indicate the direction of change in the predicted C2C values; detailed quantification of effect sizes and investigation of scene-specific reasons require additional studies and lie beyond the scope of this study.

\paragraph{Permutation Feature Importance Analysis} \label{Permutation Feature Importance}

To further compare RF and XGBoost under the same criterion, a model-agnostic measure is required. Permutation importance under RMSE provides such a baseline: each feature is shuffled on the held-out fold, and the increase in RMSE is recorded. This protocol treats both models identically and reveals differences in how they exploit features.

\begin{figure}[htb!]
  \centering
  \includegraphics[width=1.0\linewidth]{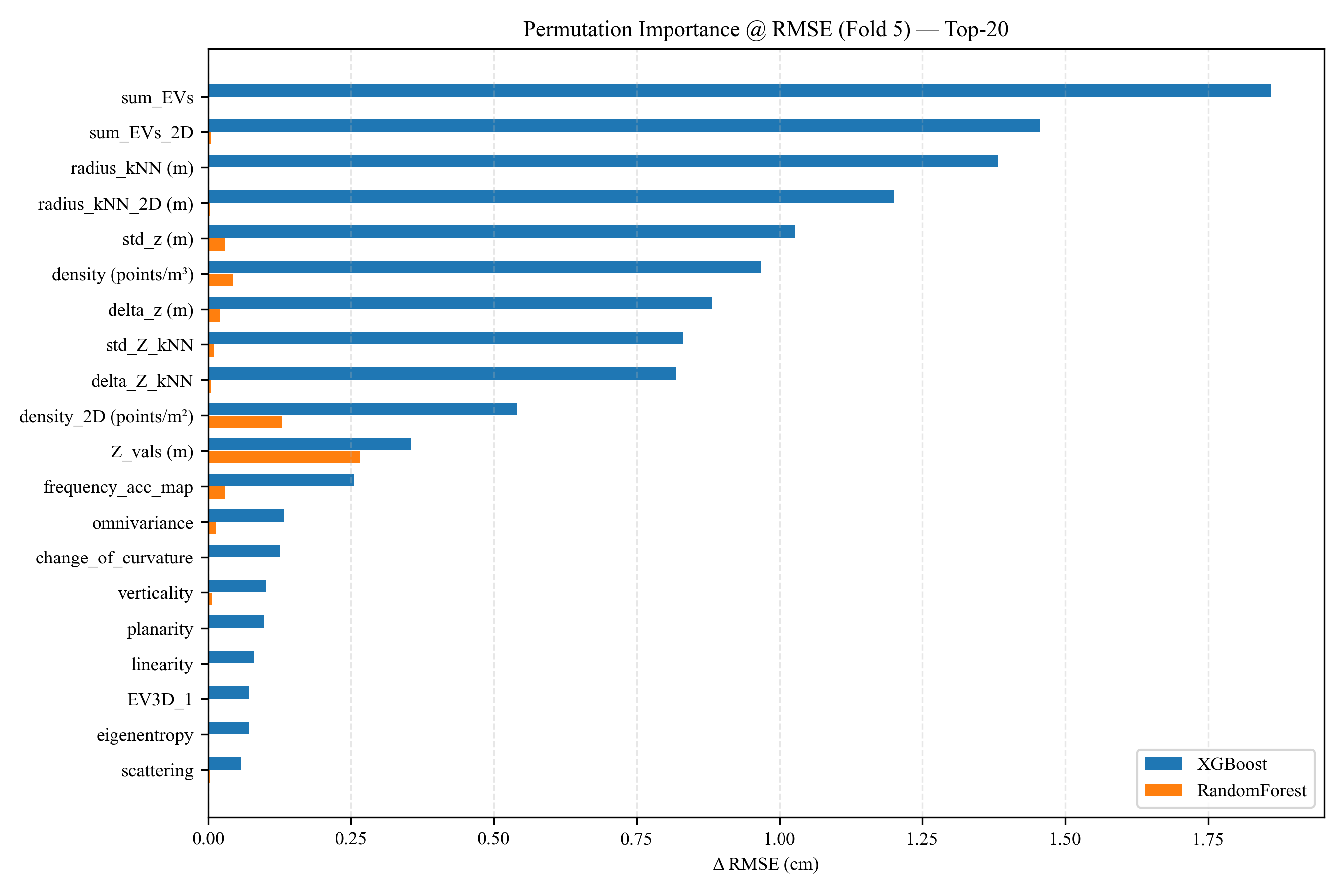}
  \caption{Permutation feature importance analysis on RF and XGBoost.}
  \label{fig:perm-compare}
\end{figure}

As seen in Figure~\ref{fig:perm-compare}, XGBoost shows large $\Delta$RMSE for several groups of features, including eigenvalue- and density-based ones. This indicates broad and effective use of complementary information. RF exhibits smaller $\Delta$RMSE overall, with noticeable reliance on $density\_2D$ and $Z\_vals$ and limited sensitivity to other features. The XGBoost permutation ranking is also consistent with the SHAP analysis, which also highlights height, density, and roughness as dominant factors. Differences between the two models are linked to their learning mechanisms: gradient boosting with regularization tends to place greater weight on leading features, whereas bagging with depth limits tends to average weaker trees and is comparatively robust to the influence of individual features.

\section{Discussion}\label{Discussion}

\paragraph{Advantages}\label{Advantages}

Learning-based modeling has opened up a new research frontier and offers three key advantages: First, it effectively reduces reliance on the GT, making it more practically relevant. Second, it preliminarily demonstrates that uncertainty is learnable, broadening the scope of current point cloud uncertainty research. Third, compared to rigid error propagation, it provides a more flexible, task-customizable framework, allowing data users to define error metrics and design prediction models from their perspective.

\paragraph{Challenges}\label{Challenges}

Despite the promising results given by the proposed framework, several challenges remain. The generalization ability of the framework has not yet been verified across diverse datasets and platforms, so external validation is required. The necessity of using all 26 geometric features is also unclear \cite{weinmann2013}; systematic ablation and feature pruning may reveal smaller subsets that achieve comparable or better accuracy with improved efficiency. In addition, the observed $R^2$ is modest, which calls for further investigation into potential causes such as label noise, feature distribution, and protocol choices.

\section{Conclusion}\label{Conclusion}

This study aims to predict point-level uncertainty in MLS point clouds using machine learning. It introduces a learning-based framework that integrates optimal neighborhood estimation with geometric feature extraction and validates the framework on a real-world dataset. The results indicate that point clouds' uncertainty is learnable and XGBoost achieves accuracy comparable to RF while providing higher efficiency. Additionally, feature-importance analyses offer initial evidence of relationships between geometric features and the C2C distance. Overall, the study explores a novel learning-based paradigm for point clouds uncertainty modeling.

For the future, it is imperative to explore the generalizability and transferability of learning-based modeling. Additionally, investigating other advanced learning strategies, such as symbolic regression \cite{Cranmer2023}, physics-informed learning, or Graph Neural Network (GNN), may yield valuable insights.

\section*{Acknowledgment}
This research was funded by TUM Georg Nemetschek Institute of Artificial Intelligence for the Built World, project "NERF2BIM”, PI Christoph Holst.

{
	\begin{spacing}{1.17}
		\normalsize
		\bibliography{main} 
	\end{spacing}
}

\end{document}